\documentclass[a4paper]{article}

\usepackage{INTERSPEECH2020}

\title{Multilingual Speech Recognition for Low-Resource Indian Languages using Multi-Task conformer}
\name{Krishna D N}
\address{
  Freshworks Inc.
  }
\email{krishna.nanjappa@freshworks.com}

\begin{document}

\maketitle
\begin{abstract}
Transformers have recently become very popular for sequence-to-sequence applications such as machine translation and speech recognition. In this work, we propose a multi-task learning-based transformer model for low-resource multilingual speech recognition for Indian languages. Our proposed model consists of a conformer [1] encoder and two parallel transformer decoders. We use a phoneme decoder (PHN-DEC) for the phoneme recognition task and a grapheme decoder (GRP-DEC) to predict grapheme sequence. We consider the phoneme recognition task as an auxiliary task for our multi-task learning framework. We jointly optimize the network for both phoneme and grapheme recognition tasks using Joint CTC-Attention [2] training. We use a conditional decoding scheme to inject the language information into the model before predicting the grapheme sequence. Our experiments show that our proposed approach can obtain significant improvement over previous approaches [4]. We also show that our conformer-based dual-decoder approach outperforms both the transformer-based dual-decoder approach and single decoder approach. Finally, We compare monolingual ASR models with our proposed multilingual ASR approach.

\end{abstract} 

\noindent\textbf{Index Terms}: multilingual speech recognition, transformers, multi-task learning, ASR, low-resource

\section{Introduction}
Studies have shown that more than 22 main spoken languages in India, and apart from Hindi and Indian English, many other languages are considered low resources due to the scarcity of the data and the speaker population. However, many Indian languages have acoustic similarities, and using multilingual acoustic models can be a good fit for building better speech recognition models. In this work, we explore multilingual speech recognition models for Indian languages.

Recent advances in deep learning have shown significant improvements in the speech recognition field. Conventional or Hybrid speech recognition models use a combination of Gaussian Mixture Models(GMMs) with Hidden Markov Models (HMMs) or Deep neural networks (DNNs) with HMMs. Many hybrid approaches have been proposed in the past [5,6,7] for multilingual speech recognition, and these approaches use initial layers of the acoustic models as feature extractors for language-dependent layers during adaptation. [8] proposes TDNN-HMM multilingual acoustic model with monolingual language model during decoding. [9] uses multi-task DNN acoustic model, which predicts both context-dependent and context-independent targets for the multilingual speech recognition task. [10] proposed to use LF-MMI [11] based TDNN acoustic models for low resource multilingual speech recognition. Hybrid multilingual ASR models with multi-task learning have been proposed in [3, 12]. Hybrid models require components like an acoustic model, a pronunciation dictionary, and a language model. These components may become hard to build for low resource languages.

End-to-End (E2E) approaches for speech recognition have become very popular in recent years [2,13,31]. Unlike hybrid models, E2E models do not require a pronunciation dictionary or alignments to train. In the case of E2E speech recognition, acoustic models, pronunciation dictionaries, and language models are integrated into a single framework. Sequence-to-sequence [14] models are a particular class of E2E models, which use encoder and decoder framework to learn a mapping from acoustic data to text data. [15] proposed an attention-based E2E model called LAS using LSTM based encoder and decoder. [16] proposed to use the LAS model for multilingual speech recognition for 12 Indian languages, and they showed that training a single model for all the languages improves each language's performance. [17] propose using language identification as an auxiliary task for the LAS framework, and they show that it can improve speech recognition performance. Recently, Transformers [18] have state of the art performance for many Natural language processing problems [19]. The transformer models are being used in many areas of speech including speech recognition [1,20,21,22], emotion recognition [23], spoken language translation [24] etc. Transformer-based approaches for multilingual speech recognition [4, 25] have shown promising results recently.

This paper proposes a dual-decoder conformer model for multilingual speech recognition with a multi-task learning framework. Our model consists of a single conformer encoder and two transformer decoders called phoneme decoder (PHN-DEC) and grapheme decoder (GRP-DEC). The phoneme decoder predicts a sequence of phonemes, and the grapheme decoder predicts a sequence of grapheme units for a given utterance. The grapheme decoder predicts a sequence of grapheme units from a multilingual vocabulary that contains all the grapheme units for all the languages. The vocabulary is also augmented with language labels. The phoneme sequence prediction acts as an auxiliary task for the model during training. The grapheme decoder is tasked to predict the language information before generating the letter sequence, and this way, we can induce the language information to the model during training. We conduct our experiments on low-resource Indian language speech recognition data released by Microsoft and SpeechOcean.com as part of a special session on "Low resource speech recognition challenge on Indian languages in INTERSPEECH 2018". We show that our approach obtains significant improvements over the baseline models.

The organization of the paper is as follows. Section 2 explains our proposed approach in detail. In section 3, we give details of the dataset and experimental setup. Finally, section 4 describes our results.

\section{Proposed approach}
End-to-end speech models have shown great promise for speech recognition applications in recent years. In this work, we propose a dual-decoder multilingual transformer model for multilingual speech recognition. Our proposed model architecture is presented in Figure 1. It consists of a conformer Encoder and two transformer decoders. The PHN-DEC generates phoneme sequence for the given utterance, whereas GRP-DEC predicts the grapheme sequence along with language information. The phoneme recognition task helps to learn shared feature representation across tasks and improves the overall system performance. The entire network is trained in an end-end-fashion using the joint CTC-Attention objective. The loss function consists of a weighted sum of CTC loss, phoneme recognition loss, and grapheme recognition loss. We describe each part of the network in detail in the following section.

\begin{figure}[t]
  \centering
  \includegraphics[width=\linewidth]{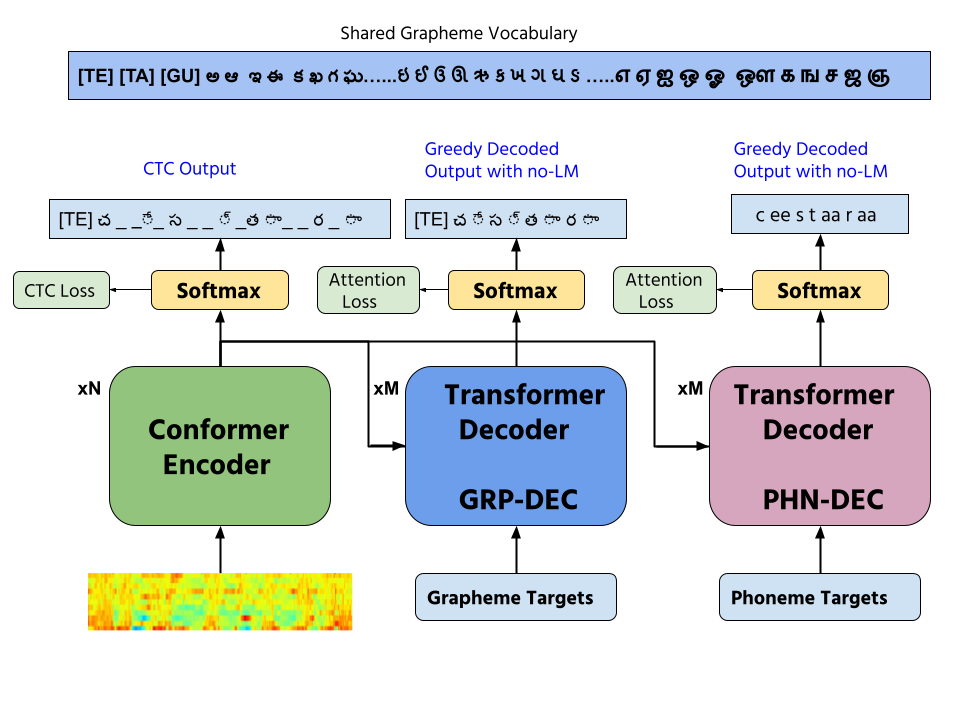}
  \caption{Proposed model architecture.}
  \label{fig:speech_production}
\end{figure}

\subsection{Conformer Encoder}
The conformer model is a variant of transformer architecture where the encoder's self-attention layers are augmented with convolution layers. The multi-head self-attention layers use a scaled dot-product attention mechanism to capture the global variations in the feature sequence. Since the convolution operation captures the local information, augmenting transformer layers with convolution layers improves the overall system performance. Our conformer encoder module is presented in Figure 2. It consists of a convolution sub-sampling layer followed by N conformer blocks. Each conformer block consists of 2 feed-forward (Linear) layers, a convolution module, and a layer normalization layer, as shown in Figure 2 (left). Processing longer feature sequences could be problematic for transformer architectures due to linear complexity in the self-attention layers. To avoid memory issues, we use an initial sub-sampling layer to reduce the input sequence length. The convolution sub-sampling reduces the frame rate by a factor of 4. The convolution module in the conformer block is responsible for learning local feature representation from feature sequences. The blocks of the convolution module are shown in Figure 2 (right). The convolution module consists of a pointwise convolution layer with an expansion factor of 2 with a GLU activation layer followed by a 1-D Depthwise convolution layer, Batch normalization, and a swish activation. 
The conformer encoder takes a sequence of acoustic features as input and transforms it through multiple transformation layers using conformer blocks to learn high-level feature representation. The conformer encoder's output is used as input to the PHN-DEC, GRP-DEC, and CTC. During decoding, we use the GRP-DEC output directly without any language model integration.

\begin{figure}[t]
  \centering
  \includegraphics[width=\linewidth]{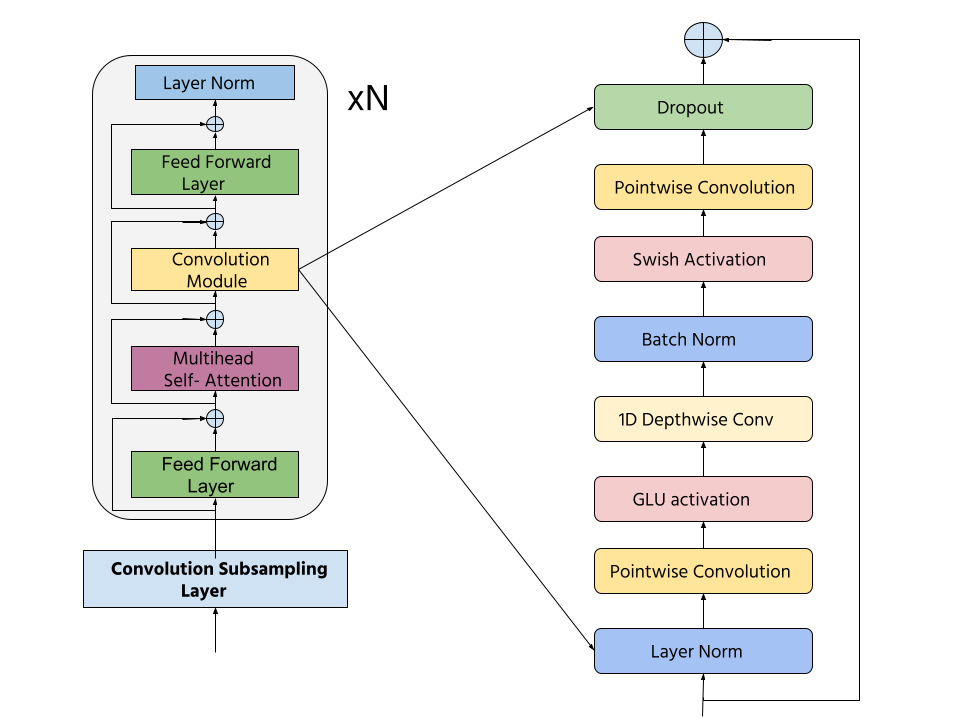}
  \caption{Conformer Encoder}
  \label{fig:speech_production}
\end{figure}

\subsection{Dual-Decoder Network}
Recent end-to-end speech recognition models consist of an encoder-decoder framework where the encoder processes the audio features and the decoder generates a sequence of output units (characters or phonemes). This paper proposes using two different decoders known as phoneme decoder (PHN-DEC) and grapheme decoder (GRP-DEC). The PHN-DEC acts as an auxiliary network to improve the overall system performance. The phoneme decoder consists of a single transformer layer, and it takes conformer encoder output and phoneme embeddings as input, as shown in Figure 1. Each decoder layer in PHN-DEC consists of feed-forward layers and multi-head cross attention layers. The output of the PHN-DEC is a sequence of phoneme corresponding to the input utterance. We use 58 phonemes from the common phoneme label set provided by IndicTTS\footnote{https://www.iitm.ac.in/donlab/tts/unified.php}. The grapheme decoder (GRP-DEC) is also a transformer-based decoder consisting of transformer layers. The GRP-DEC takes character embeddings and conformer encoder output as input to predict the sequence of grapheme units for a given input utterance. We propose using a conditional decoding scheme to force the GRP-DEC to predict the language information before predicting a sequence of grapheme units. This way, we can induce language information to the network and force the network to learn the language label of the given input utterance. The GRP-DEC vocabulary consists of all possible characters from 3 Indian languages Telugu, Tamil, and Gujarati. The vocabulary is also augmented with language labels, as shown in Figure 1.

\subsection{Multi-task Learning}
Our model uses three different loss functions, which are described below.
We use the conformer encoder output logits to the CTC layer to compute the CTC loss \boldsymbol{$\mathcal{L}_{ctc}$}. The CTC layer predicts the grapheme sequence for a given utterance. The phoneme recognition decoder PHN-DEC takes conformer encoder output and phoneme embeddings as input and generates a sequence of phoneme labels. We denote the loss calculated for the phoneme recognition task as \boldsymbol{$\mathcal{L}_{pr}$}. In the multi-task learning framework, we can consider both phoneme recognition loss and CTC loss as auxiliary losses.
The grapheme recognition decoder (GRP-DEC) is responsible for predicting the grapheme sequence for the given input audio. It takes conformer encoder output and character embeddings as input and generates a sequence of grapheme units along with the language label. We denote the loss computed by GRP-DEC as \boldsymbol{$\mathcal{L}_{gr}$}. During training, we compute the weighted sum of the 3 losses \boldsymbol{$\mathcal{L}_{ctc}$}, \boldsymbol{$\mathcal{L}_{pr}$}, and \boldsymbol{$\mathcal{L}_{gr}$} to jointly optimize the the network for speech recognition task. The multi-task loss function is given by:

\begin{equation}
  \begin{aligned}
    \boldsymbol{\mathcal{L}} = \lambda  \boldsymbol{\mathcal{L}_{ctc}} + (1-\lambda) \boldsymbol{\mathcal{L}_{gr}} + \alpha \boldsymbol{\mathcal{L}_{pr}}
 \end{aligned}
\end{equation}

Where $\lambda$ and $\alpha$ are considered as the hyperparameters of the model. The $\alpha$ parameter is called as phoneme interpolation co-efficient.

\subsection{Conditional Decoding}
Multilingual speech recognition involves training an acoustic model which learns to transcribe audios from multiple languages. Sometimes two or more languages will have similar acoustic units, which is very common in many Indian languages. During training, the model should be provided with the language information either to the encoder or decoder. Previous works [4] have investigated many ways of inducing language information including, one-hot encoding and language embedding-based techniques. There are various ways to induce language information to the model, but we investigate the condition decoding scheme in this work. The conditional decoding objective forces the GRP-DEC to predict the language label before generating the character sequence. Since the GRP-DEC decoder is autoregressive, every predicted character will be conditioned on the predicted language label. This technique helps the model to learn language identification capability.

\section{Dataset and Experiments}
We conduct our experiments on the data set released by Microsoft and SpeechOcean.com as part of a special session on “Low resource speech recognition challenge on Indian languages in INTERSPEECH 2018”. The dataset includes data for three Indian languages - Telugu, Tamil, and Gujarati. All the audio files are sampled at 16KHz, and the transcriptions contain UTF-8 text for every language. The dataset also comes with a lexicon for each language, and the lexicons are built with common label set phonemes\footnote{https://www.iitm.ac.in/donlab/tts/cls.php}. The dataset contains train, development, and evaluation sets for each language. Unfortunately, the transcripts for the evaluation sets are not made public. Each language contains 40hrs of training data and 5hrs dev data. We use 3hrs of randomly selected audio files from the train set from each language as a validation set during training. We report all our results on the development set and compare our model performance with previous research published on the same set. 

The conformer encoder contains 12 conformer blocks with two initial convolution layers for sub-sampling. Each conformer block contains eight attention heads with 512 units for the attention head and 2048 hidden units for the pointwise feed-forward layer. It also uses the positional encoding layer to learn positional information of the speech frames. Both PHN-DEC and GRP-DEC use single transformer layers with eight attention heads and 2048 hidden units for the pointwise feed-forward layer. The vocabulary of PHN-DEC consists of 58 phonemes excluding $<$unk$>$, $<$blank$>$,$<$space$>$ and $<$sos/eos$>$ tags. The vocabulary of GRP-DEC consists of 178 graphemes from all the languages combined without extra tags. We augment the vocabulary $<$unk$>$, $<$blank$>$,$<$space$>$, $<$sos/eos$>$ and three language labels [TE] (Telugu),[TA] (Tamil), and [GU] (Gujarati). Our final grapheme vocabulary contains 185 tags in total. Our model hyperparameters $\lambda$ is set to 0.3, and we vary the value of phoneme interpolation co-efficient $\alpha$ from 0.2 to 1.0 in steps of 0.1. We use 40-dimensional Mel-filterbank features extracted with a 25ms frame window, with 10ms frameshift. The training dataset is augmented with speech perturbation and SpecAugment [27]. We use Adam optimizer [28] with an initial learning rate of 0.001, and the learning rate is decayed using a learning rate scheduler based on validation loss. We train our model up to 30 epochs with a batch size of 20.

\section{Results}
In this section, we describe our experimental results for both monolingual and multilingual systems. We also show the effectiveness of the conformer Encoder with a dual-decoder model and compare it with transformer-based approaches.

\subsection{Comparison between model architectures}
In this section, we compare our approach with previously published multilingual speech recognition models [4].  The word error rate on the development set for various models is given in Table 1. The system \textit{BiLSTM} contains a four-layer BiLSTM layers in the encoder and 1 BiLSTM layer in the decoder. The \textit{Transformer} model uses a 12 layer transformer encoder and one layer transformer layer in the decoder. A similar configuration is used in \textit{Transformer}+Embeds, except the language information is provided as embeddings. A multilingual acoustic model can be adapted again for specific language using re-training. \textit{Transformer}+Embeds+retrain represents the retrained language-specific models. On the other hand, \textit{Conformer}+GRP+PHN is our proposed conformer-based E2E model. It uses 12 conformer blocks in the encoder, one transformer block in the decoder, and one transformer block in PHN-DEC. Similarly, \textit{Conformer}+GRP (ours) uses 12 conformer blocks in the encoder and one transformer block in GRP-DEC (no PHN-DEC). We compare our final model results \textit{Conformer}+GRP+PHN (last row Table1/2) with the previous best approach \textit{Transformer}+Embeds+retrain. From Table 1, we can see that we obtain a relative WER reduction of 23\% for Gujarati, 26\% for Tamil, and 27\% for Telugu. Results from  Table 1 and Table 2 shows that our proposed system outperforms the previous best approach by a large margin in terms of both WER and CER.

\begin{table}[!htbp]
  \centering
  \caption{WER comparison between previous approaches and our approach. Bold indicates the best WER (lower the better)}
  \label{tab:tasks}
    \begin{tabular}{l|c|c|c}
     \hline
      \textbf{System} & \textbf{GU} & \textbf{TA} & \textbf{TE}\\
      \hline
      \textit{BiLSTM}     & 40.00 & 39.60 & 42.70 \\
      \textit{Transformer}     & 30.90 & 34.70 & 36.40 \\
      \textit{Transformer}+Embeds [4]  & 30.00 & 33.90 & 35.40 \\
      \textit{Transformer}+Embeds+retrain [4]  & 29.20 & 33.20 & 34.80 \\
      \textit{Conformer}+GRP  & 23.92 & 27.36 & 27.83 \\
      \textit{Conformer}+GRP+PHN  & \textbf{22.38} & \textbf{24.54} & \textbf{25.28} \\
      \hline
    \end{tabular}
\end{table}

\begin{table}[!htbp]
  \centering
  \caption{CER comparison between previous approaches and our approach}
  \label{tab:tasks}
    \begin{tabular}{l|c|c|c}
     \hline
      \textbf{System} & \textbf{GU} & \textbf{TA} & \textbf{TE}\\
      \hline
      \textit{BiLSTM} [4]     & 16.10 & 9.70 & 12.0 \\
      \textit{Transformer} [4]     & 9.10 & 7.70 & 8.50 \\
      \textit{Transformer}+Embeds [4]  & 8.60 & 7.30 & 8.10 \\
      \textit{Transformer}+Embeds+retrain [4]  & 8.30 & 7.10 & 7.70 \\
      \textit{Conformer}+GRP (ours)  & 6.98 & 5.73 & 6.20 \\
      \textit{Conformer}+GRP+PHN (ours)  & \textbf{6.63} & \textbf{5.62} & \textbf{6.18} \\
      \hline
    \end{tabular}
\end{table}

\subsection{Dual-Decoder and Conformer Encoder}
In this section, we present the effectiveness of our dual-decoder multi-task learning approach and the conformer encoder. To compare the effectiveness of the encoders, we train four different models using both transformer encoder and conformer encoder, as shown in Table 3. To compare the ability of our dual-decoder approach (multi-task learning) over a single decoder approach, we build a model with a conformer encoder with only GRP-DEC (no phoneme decoder), and we call the model \textit{Conformer}+GRP (Table 3) and compare it with \textit{Conformer}+GRP+PHN which has both PHN-DEC and GRP-DEC. The word error rate on the development set for each of the models is given in Table 3. From Table 3, we can see that the conformer encoder models obtain more than a 25\% average reduction in WER compared to transformer-based approaches. From Table 3, we can also see that the multi-task learning framework can bring a significant improvement to model performance, and this is because the model learns robust and shared feature representations when it is tasked to solve multiple objectives at the same time.

\begin{table}[!htbp]
  \centering
  \caption{WER comparison between different encoders and multi-task learning. Bold indicates the best WER (lower the better)}
  \label{tab:tasks}
    \begin{tabular}{l|c|c|c}
     \hline
      \textbf{System} & \textbf{GU} & \textbf{TA} & \textbf{TE}\\
      \hline
      \textit{Transformer}+GRP & 31.01 & 35.12 & 37.79   \\
      \textit{Transformer}+GRP+PHN & 29.23 & 33.89 & 36.15 \\
      \textit{Conformer}+GRP  & 23.92 & 27.36 & 27.83 \\
      \textit{Conformer}+GRP+PHN  & \textbf{22.38} & \textbf{24.54} & \textbf{25.28} \\
      \hline
    \end{tabular}
\end{table}

\subsection{Effect of phoneme interpolation co-efficient}
In this section, we describe the role of the interpolation coefficient $\alpha$ on the system performance. We set up an experiment where we vary the value of $\alpha$ from 0 to 1.0 in steps of 0.1. We train our model for each of the values of $\alpha$ and observe both WER and CER. We use 12 conformer layers in the encoder and one transformer layer for both PHN-DEC and GRP-DEC. We keep the same model architecture for all the values of $\alpha$. The plot of WER/CER w.r.to the value of the interpolation coefficient $\alpha$ is shown in Figure 3. It can be observed that at $\alpha$ 0.6, the model obtains the best WER and CER.

\begin{figure}[t]
  \centering
  \includegraphics[width=\linewidth]{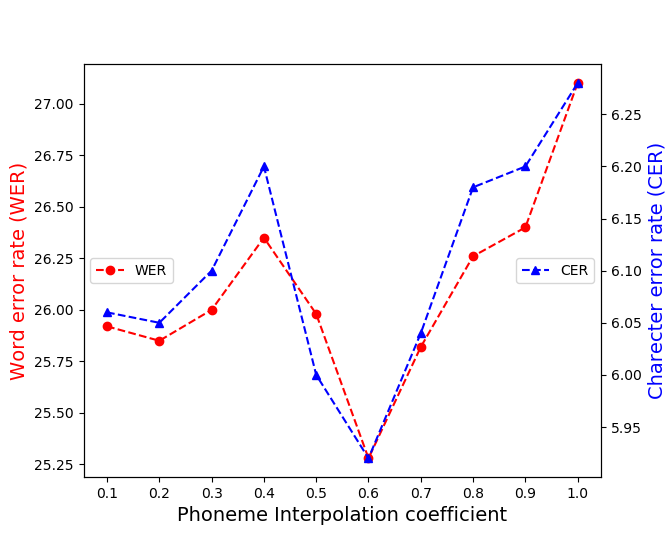}
  \caption{Effect of phoneme interpolation coefficient $\alpha$ on WER and CER}
  \label{fig:speech_production}
\end{figure}

\subsection{Monolingual vs Multilingual systems}
In this section, we compare different neural architectures for both monolingual and multilingual settings. We use BiLSTM, transformer, and conformer models for both monolingual and multilingual models. The results are shown in Table 4. Our experiments show that our method consistently obtains better performance in both monolingual and multilingual settings. From Table 4, we can see that the multilingual models gain significant improvement over monolingual counterparts due to the shared feature representation across different languages. Since many Indian languages share acoustic similarity across languages, sharing acoustic models helps to learn shared feature representation across different languages.

\begin{table}[!htbp]
  \centering
  \caption{WER comparison between monolingual multilingual models.}
  \label{tab:tasks}
    \begin{tabular}{l|c|c|c}
     \hline
      \textbf{System} & \textbf{GU} & \textbf{TA} & \textbf{TE} \\
      \hline
      \textbf{Monolingual} & & &\\
      \textit{BiLSTM} [4]     & 45.3 & 39.6 & 43.9 \\
      \textit{Transformer} [4]     & 31.90 & 34.10 & 36.00 \\
      \textit{Conformer}+GRP (ours)  & 27.83 & 31.00 & 33.90 \\
      \textit{Conformer}+GRP+PHN (ours) & \textbf{26.68} & \textbf{28.10} & \textbf{29.92}\\
      \hline
      \textbf{Multilingual} & & & \\
      \textit{BiLSTM} [4]     & 40.00 & 39.60 & 42.70 \\
      \textit{Transformer} [4]     & 30.90 & 34.70 & 36.40 \\
      \textit{Transformer}+Embeds+retrain [4]  & 29.20 & 33.20 & 34.80 \\
      \textit{Conformer}+GRP  & 23.92 & 27.36 & 27.83 \\
      \textit{Conformer}+GRP+PHN  & \textbf{22.38} & \textbf{24.54} & \textbf{25.28} \\
      
      \hline
    \end{tabular}
\end{table}

\section{Conclusions}
Multilingual speech recognition is one of the challenging problems in speech processing. In this work, we propose a dual-decoder approach for multilingual speech recognition tasks for three Indian languages Gujarati, Tamil, and Telugu. We propose to use a conformer encoder with two parallel transformer decoders. We use the phoneme recognition task as an auxiliary task to learn shared feature representation in the multi-task learning framework. We show that our dual-decoder approach obtains more than a 25\% average relative reduction in WER compared to the previous transformer-based single decoder approach. We also show that our proposed conformer encoder systems obtain a significant reduction over the transformer encoder systems. 

\bibliographystyle{IEEEtran}

\end{document}